\documentclass[3p,times,procedia]{elsarticle}
\usepackage{ecrc}
\usepackage[bookmarks=false]{hyperref}
    \hypersetup{colorlinks,
      linkcolor=blue,
      citecolor=blue,
      urlcolor=blue}

\volume{00}

\firstpage{1}

\journalname{Procedia Computer Science}

\runauth{J. Ignatowicz et al.}

\jid{procs}

\usepackage{amssymb}

\usepackage[figuresright]{rotating}

\usepackage{wrapfig,booktabs}

\usepackage{color, colortbl}
\definecolor{Gray}{gray}{0.9}

\usepackage[subtle]{savetrees}

\begin{document}
\begin{frontmatter}



\dochead{28th International Conference on Knowledge-Based and Intelligent Information \& Engineering Systems (KES 2024)}%

\title{Evaluation and Comparison of \\ Emotionally Evocative Image Augmentation Methods}


\author{Jan Ignatowicz\corref{cor1}} 
\author{Krzysztof Kutt}
\author{Grzegorz J. Nalepa}

\address{Jagiellonian Human-Centered AI Lab, Mark Kac Center for Complex Systems Research, Institute of Applied Computer Science, Faculty of Physics, Astronomy and Applied Computer Science, Jagiellonian University, Łojasiewicza 11, 30-348, Krak\'ow, Poland}

\begin{abstract}
Experiments in affective computing are based on stimulus datasets that, in the process of standardization, receive metadata describing which emotions each stimulus evokes. 
In this paper, we explore an approach to creating stimulus datasets for affective computing using generative adversarial networks (GANs). Traditional dataset preparation methods are costly and time consuming, prompting our investigation of alternatives. We conducted experiments with various GAN architectures, including Deep Convolutional GAN, Conditional GAN, Auxiliary Classifier GAN, Progressive Augmentation GAN, and Wasserstein GAN, alongside data augmentation and transfer learning techniques. Our findings highlight promising advances in the generation of emotionally evocative synthetic images, suggesting significant potential for future research and improvements in this domain.
\end{abstract}

\cortext[cor1]{Corresponding author.}
\end{frontmatter}

\email{jan.ignatowicz@doctoral.uj.edu.pl, krzysztof.kutt@uj.edu.pl, gjn@gjn.re}




\section{Introduction and Motivation}
\label{sec:intro}


Affective computing (AfC)---an interdisciplinary area of research on emotions involving their recognition, processing, and simulation in computer systems~\cite{picard1997affective}---is based on, inter alia, carefully planned experiments in which participants are exposed to stimuli that evoke specific emotions~\cite{kkt2022nature}.
The reactions are then collected in an appropriate manner, recorded, and finally, relevant emotion models are developed based on them.


Among the most common types of stimuli are images, usually depicting scenes, faces, or objects intended to evoke specific emotions (see Fig.~\ref{fig:oasis_images}).
These are grouped in datasets, which usually contain dozens to more than a thousand stimuli, along with metadata that specifies the emotion the image evokes~\cite{carretie2019emomadrid}.
However, since a single study may use several hundred stimuli, the few existing datasets containing affective images contain too few stimuli relative to research needs.
A participant taking part in a few studies has a high chance of encountering images that they are already familiar with, making their responses weaker due to habituation effect~\cite{carretie2019emomadrid}.


The solution to the problem would be to create new datasets or expand existing ones, but in a ``classical'' approach this is very costly, due to the need to present each new stimulus to a sufficiently large study group to provide standardized metadata (e.g., \cite{lang2008iaps,marchewka2014naps}).
In this paper, we want to present a different solution, based on the use of generative models to automatically learn the natural features of a dataset. 
To the best of our knowledge, no one has previously reported an attempt to create a dataset of affective images in this manner.


The task of generative models is to produce new synthetic samples with features that look like the features of the real elements of the training set~\cite{foster2019generative}.
The task of generative models is therefore unsupervised learning, in which the model tries to capture the properties of the dataset and represent them in a probabilistic rather than deterministic manner, as it must be able to generate many new samples.
To adapt this approach to the needs of AfC, we divided the training set into subsets containing images that evoke distinct emotions and prepared an image label generative model for each emotion so selected.


The paper is organized as follows.
In Sect.~\ref{sec:generative}, we outline generative models considered in the study.
Then, in Sect.~\ref{sec:datasets}, datasets used in the research are summarized.
The data processing and models training are described in Sect.~\ref{sec:training}.
The evaluation is reported in Sect.~\ref{sec:results}. In Sect.~\ref{sec:dalle} DALL-E usage is presented as fulfillment of affective generation task and the paper is concluded in Sect.~\ref{sec:summary}.


\section{Generative Models}
\label{sec:generative}

One of the most important attempts to handle with generative modeling problem was presented in ~\cite{goodfellow-gans}.
Ian J. Goodfellow introduced a new machine learning framework named Generative Adversarial Networks (GANs).
The architecture of this framework is constructed with two distinct models, further called generator and discriminator.
The generator is supposed to create new images similar to the training images while the discriminator has to claim whether achieved input image is real or fake.

The GAN training is performed on the minimax algorithm working according to the Nash equilibrium.
Both networks (discriminative and generative) learn in the best possible way for them.
With each learning epoch the generator tries to produce better fake images.
Meanwhile the discriminator is still learning which examples are real or fake and tries to find them out every time.
The equilibrium of this minimax game is obtained when the generator generates perfect fake images and the discriminator cannot judge which image is real and which is fake.
It leads to the situation when discriminator always gives his response at 50\% confidence level.



The research utilized various GAN architectures, notably the Deep Convolutional GAN (DCGAN)~\cite{dcgan} which employs convolution layers for processing. Conditional GANs (cGAN)~\cite{cgan} and Auxiliary Classifier GANs (ACGAN)~\cite{acgan} were used for image generation with categorical emotion conditions, improving data generation for imbalanced datasets by embedding additional label information into both generator and discriminator networks. Progressive Augmentation GANs (PAGANs)~\cite{pagan} address discriminator stability by gradually increasing task difficulty, acting as a regularization technique. Lastly, Wasserstein GANs (WGAN)~\cite{wgan} and its variant with Gradient Penalty~\cite{wgan-gp} focus on modifying the cost function to enhance training effectiveness, demonstrating an alternate approach to generator training that lessens the importance of network balance.

Dataset augmentation is crucial for machine learning, especially in computer vision, to prevent overfitting due to small datasets. Numerous methods~\cite{data-augmentation} exist to artificially expand datasets, significantly enhancing model performance.

Transfer Learning (TL) repurposes knowledge from one domain to another, crucial for tasks with limited data. It includes using pretrained ConvNets as feature extractors or fine-tuning them on new datasets, aiding in overcoming dataset size limitations, particularly when pretrained on extensive collections like ImageNet~\cite{imagenet}.

Pretrained models, such as BigGAN~\cite{biggan} and StyleGAN~\cite{style-gan}, provide a foundation for generating quality images. Despite the rise of TL, pretrained models, mainly for classification, remain pivotal. They can be retrained with affective datasets to augment data with new, label-linked images.

Effective GAN evaluation metrics should balance~\cite{metrics},~\cite{metrics2} fidelity, diversity, and controllability of samples, align with human judgment, and maintain low complexity. Key metrics include the Inception Score (IS)~\cite{is}, measuring quality and diversity; Fréchet Inception Distance (FID)~\cite{fid}, assessing similarity to real images; and Kernel Inception Distance (KID)~\cite{kid}, a variant of FID emphasizing statistical similarity.


\section{Datasets}
\label{sec:datasets}

\begin{figure}[htp!]
    \centering
    [a\label{fig:valence_arousal}]{
      \includegraphics[width=0.56\textwidth]{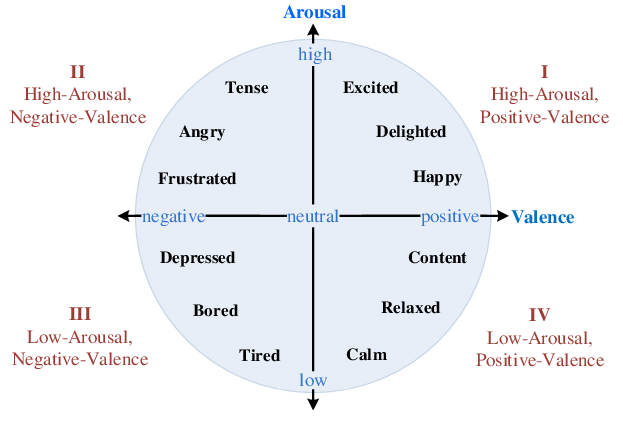}%
    }\hfil
    [b\label{fig:oasis_images}]{
      \includegraphics[width=0.36\textwidth]{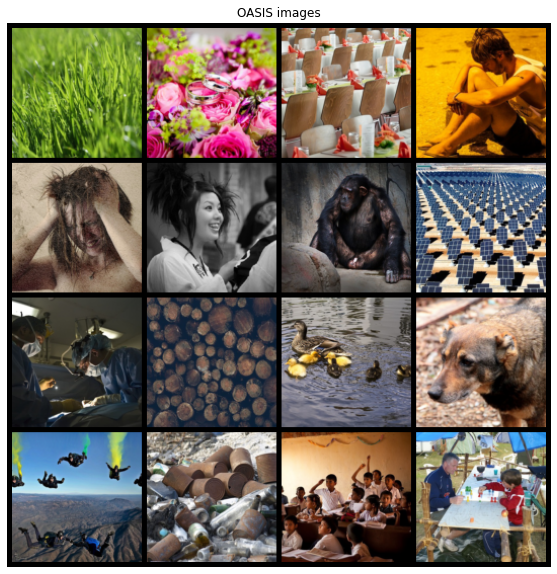}%
    }
    \caption{[a] Russel's circumplex model of emotions~\cite{val-aro-diagram}, [b] Pseudo-randomly selected OASIS images (based on~\cite{oasis})}
    \label{fig:valence_arousal_oasis_images}
\end{figure}

This study employed six datasets of affective images, encompassing a wide array of emotions and scenes. The International Affective Picture System (IAPS)~\cite{lang2008iaps} includes a broad spectrum of images such as animals, people, everyday objects, and landscapes. The Geneva Affective Picture Database (GAPED)~\cite{gaped} focuses on images eliciting negative, neutral, and positive emotions through themes like moral violations, animal mistreatment, and more uplifting images of babies and landscapes. The Nencki Affective Picture System (NAPS)~\cite{marchewka2014naps} provides high-quality photographs across five categories: people, faces, animals, objects, and landscapes. The Set of Fear Including Pictures (SFIP)~\cite{michalowski2017sfip} offers images specifically designed to induce fear, covering categories such as social exposure and small animals, filling a niche gap in affective imagery. The Open Affective Standardized Image Set (OASIS)~\cite{oasis} and EmoMadrid~\cite{carretie2019emomadrid} both supply open-access images sourced from the Internet across categories like landscapes, animals, and objects.

Collectively, these datasets contribute 5866 unique images to the research. Each image is annotated according to Russel's circumplex model of emotions, with two key dimensions: valence (ranging from pleasant to unpleasant) and arousal (ranging from calm to excitement), enabling the study of complex emotions, using just two dimensions~\cite{dzedzickis2020sensors} (see Fig.~\ref{fig:valence_arousal_oasis_images}[a]). Despite the original datasets using various scales for these dimensions, they were normalized to a consistent range of $[-1, 1]$ for this analysis.

5866 images were collected in total (for sample images, see Fig.~\ref{fig:valence_arousal_oasis_images}[b]), along with the ratings of valence and arousal (see Fig.~\ref{fig:all_data_valence_arousal_and_classified}[a]).
The amounts of images per quarter is shown in the Tab.~\ref{table:quarter_amounts}. The representations of each quarter are not equal, indeed they are quantitatively distant from each other. The biggest difference is between the third and fourth quarter. All this is due to too little data, especially the third quarter has significant gaps in its representation. Most of the data provided in the third quarter was contributed by the SFIP collection.

\begin{table}[htp!]
    \caption{Number of images divided by quarters (cf. Fig.~\ref{fig:all_data_valence_arousal_and_classified}[a])}
    \centering
    \begin{tabular}{|c|c|c|c|c|}
    \hline
        \textbf{Dataset}& \textbf{Quarter I}& \textbf{Quarter II}& \textbf{Quarter III}& \textbf{Quarter IV} \\ \hline
        EmoMadrid & 306 & 237 & 19 & 251 \\ \hline
        GAPED & 22 & 360 & 100 & 248 \\ \hline
        IAPS & 229 & 314 & 219 & 420 \\ \hline
        NAPS & 192 & 480 & 45 & 639 \\ \hline
        OASIS & 198 & 145 & 162 & 395 \\ \hline
        SFIP & 0 & 17 & 326 & 542 \\ \hline
        \rowcolor{Gray} ALL & 946 & 1553 & 871 & 2495 \\ \hline 
    \end{tabular}
    \label{table:quarter_amounts}
\end{table}


\section{Generation of Affective Pictures}
\label{sec:training}

This section outlines the ablation study focused on data preparation, model implementation, and leveraging publicly available resources for dataset augmentation. Implementations draw on state-of-the-art techniques, tailored to specific data needs with enhancements for shared architectural elements. The foundation is the DCGAN architecture for image generation, expanded with conditional elements for class-specific image creation. Model variations stem from their unique development requirements, with a uniform training approach across all models.




\subsection{Preprocessing}

The prepared Affective Dataset contains all the images gathered in one place and all the labels juxtapositioned in the same schema along with connection two-dimensional labels to the images. 

\begin{figure}[t]
    \centering
    [a\label{fig:all_data_valence_arousal}]{
      \includegraphics[width=0.47\textwidth]{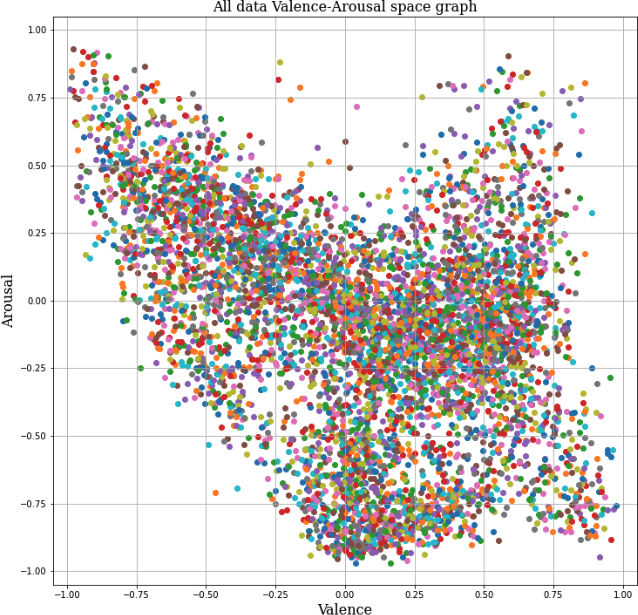}%
    }\hfil
    [b\label{fig:all_data_valence_arousal_classified}]{
      \includegraphics[width=0.47\textwidth]{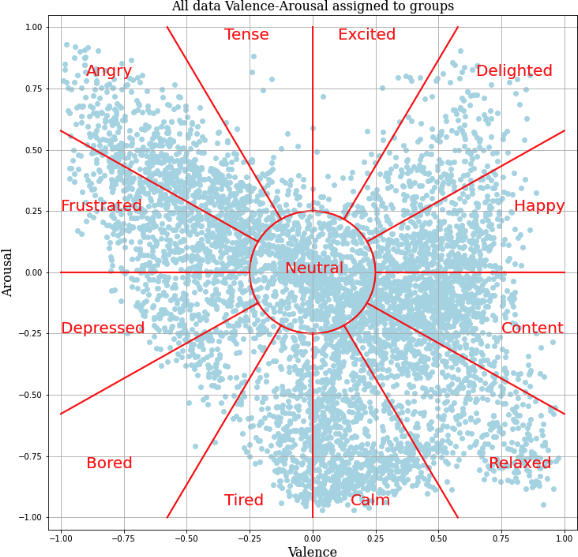}%
    }
    \caption{[a] Normalized ratings for images from all datasets, [b] Valence-arousal space divided into 13 categories (cf. Fig.~\ref{fig:all_data_valence_arousal_and_classified}[a])}
    \label{fig:all_data_valence_arousal_and_classified}
\end{figure}

For dimension reduction reason and in order to apply labels to images, the two-dimensional valence-arousal space was divided into 13 classes (Fig.~\ref{fig:all_data_valence_arousal_and_classified}[b]) to reflect model depicted on Fig.~\ref{fig:valence_arousal_oasis_images}[a].
In summary, the Table \ref{table:categories_amounts} presents the amounts of data consisting of each category. By comparing this data, it is again confirmed, that there exists a problem with representations of some classes. The most unfilled with data spaces are for categories: Tense (50 images) and Excited (36 images). There can be also observed that spread of data representation is very wide; for Neutral category there are 1036 representing images whereas for Excited category there are only 36 representing images. 

\begin{table}[t]
    \centering
    \caption{Amounts of each categories data representations}
    \begin{tabular}{|c|c|c|c|c|c|c|c|}
    \hline
        \textbf{Category} & Angry & Bored & Calm & Content & Delighted & Depressed & Excited \\ \hline
        \textbf{Amount} & 628 & 124 & 776 & 717 & 249 & 221 & 36 \\ \hline

        \textbf{Category} &  Frustrated & Happy & Neutral & Relaxed & Tense & Tired & \\ \hline
        \textbf{Amount} & 603 & 461 & 1036 & 627 & 50 & 338 & \\ \hline
    \end{tabular}
    \label{table:categories_amounts}
\end{table}

\subsection{Generative models}
The study incorporated data augmentation to enhance model performance, utilizing the Pillow library to expand an Affective Dataset. Techniques included image enhancement with detail and edge filters, brightness adjustment (lightening by a factor of 1.2 and darkening by 0.9), and rotation at 90, 180, and 270 degrees. Each augmented image retained the original's category, leading to a sevenfold increase per image, resulting in a total of 46 928 images in the augmented dataset.

A standard Deep Convolutional GAN architecture served as the benchmark, with the discriminator network evaluated in three configurations to determine optimal performance: Standard DCGAN with Batch Normalization, DCGAN with Batch Normalization and Dropout layers (DCGAN D.) and DCGAN using Spectral Normalization from Progressive Augmentation GAN (DCGAN SN).
All three variations shared hyper-parameters, employing GAN hacks like differing learning rates for the generator and discriminator and setting the real label at 90\% of truth.
The CGAN models resemble DCGAN with class embeddings and have been adjusted similarly with dropout layers and Spectral Normalization. Hyper-parameters align with DCGAN, adding class numbers for CIFAR-10 and Affective Dataset.

PAGAN and WGAN GP models build on DCGAN's architecture, sharing the same generator structure. WGAN GP differs by omitting the sigmoid function in the discriminator. PAGAN introduces extra channels to the discriminator's first convolutional layer based on the KID score, with label smoothing omitted to preserve augmentation effects. WGAN GP incorporates gradient penalty in its loss function, maintaining DCGAN's learning strategies.

Transfer learning involved selecting prominent models from Pytorch and HuggingFace, fine-tuning them with the Affective Dataset. Chosen models include ResNet-18~\cite{resnet}, ResNet-152, VGG19~\cite{vgg}, and EfficientNet\_b7~\cite{efficientnet}, with uniform training loops and specified hyper-parameters. The Affective Dataset was split into training and validation sets (80/20 ratio).

Fine-tuning the BigGAN~\cite{biggan} generator posed challenges due to the absence of a pretrained discriminator. Options include creating a discriminator or leveraging another pretrained model for classification. The former necessitates training from scratch, potentially pretraining on ImageNet, while the latter depends on successful classification task performance to replace the discriminator in fine-tuning BigGAN.




\section{Evaluation}
\label{sec:results}

\subsection{Experiments summary}

To validate the constructed models' correctness, additional experiments were conducted using the CIFAR-10~\cite{cifar10} dataset. This preliminary testing aimed to ensure the models were well-designed and capable of generating images resembling those in the real dataset, given GANs' susceptibility to issues like mode-collapse. Three discriminator variants (employing batch normalization, dropout layers, and spectral normalization) were tested across three datasets: CIFAR-10, the Affective Dataset, and the Augmented Affective Dataset, using FID and KID scores calculated every five epochs from image batches of 200.

Overall, 36 tests were performed on the generative models, each trained for 100 epochs, determined as sufficient for stabilizing image generation based on FID and KID scores. Additionally, eight tests focused on fine-tuning pretrained models for classifying affective images, conducted over 25 epochs, though optimal performance generally emerged by the fifth epoch.

The underwhelming results from fine-tuning pretrained models indicated their unsuitability as discriminators. In attempts to fine-tune the BigGAN model with a DCGAN-like discriminator, early tests showed the generator produced flawed examples within just a few epochs, leading to the discontinuation of tests with pretrained generative models.

\subsection{Generation Results}

\begin{table}[htp!]
    \centering
    \caption{Juxtaposition of best FID and KID scores for all trained models}
    {\begin{tabular}{|c|c|c|c|}
    \hline
        \textbf{Model} & \textbf{Dataset} & \textbf{FID score} & \textbf{KID score} \\ \hline
        DCGAN & CIFAR-10 & 119.9306 & 0.003294 \\ \cline{2-4}
         & AFFECTIVE & 226.8570 & 0.008311 \\ \cline{2-4}
         & A. AFFECTIVE & 173.2516 & 0.000991 \\ \hline
        DCGAN D. & CIFAR-10 & 108.3639 & 0.000300 \\ \cline{2-4}
         & AFFECTIVE & 198.9459 & 0.004590 \\ \cline{2-4}
         & A. AFFECTIVE & 168.1698 & 0.001028 \\ \hline
        DCGAN SN. & CIFAR-10 & 152.7030 & 0.005971 \\ \cline{2-4}
         & AFFECTIVE & 372.8274 & 0.025878 \\ \cline{2-4}
         & A. AFFECTIVE & 388.6158 & 0.027513 \\ \hline
        CGAN & CIFAR-10 & 159.3475 & 0.006392 \\ \cline{2-4}
         & AFFECTIVE & 232.0317 & 0.009073 \\ \cline{2-4}
         & A. AFFECTIVE & 193.2394 & 0.002538 \\ \hline
        CGAN D. & CIFAR-10 & 142.2123 & 0.004740 \\ \cline{2-4}
         & AFFECTIVE & 232.2346 & 0.009290 \\ \cline{2-4}
         & A. AFFECTIVE & 187.0331 & 0.001799 \\ \hline
        CGAN SN. & CIFAR-10 & 146.8453 & 0.005457 \\ \cline{2-4}
         & AFFECTIVE & 242.1963 & 0.010768 \\ \cline{2-4}
         & A. AFFECTIVE & 189.2623 & 0.001932 \\ \hline
        PAGAN & CIFAR-10 & 140.8431 & 0.003949 \\ \cline{2-4}
         & AFFECTIVE & 250.4171 & 0.011332 \\ \cline{2-4}
         & A. AFFECTIVE & 199.4383 & 0.008943 \\ \hline
        PAGAN D. & CIFAR-10 & 115.1276 & 0.002150 \\ \cline{2-4}
         & AFFECTIVE & 226.3566 & 0.008486 \\ \cline{2-4}
         & A. AFFECTIVE & 182.8493 & 0.003149 \\ \hline
        PAGAN SN. & CIFAR-10 & 115.8432 & 0.001387 \\ \cline{2-4}
         & AFFECTIVE & 227.9139 & 0.009626 \\ \cline{2-4}
         & A. AFFECTIVE & 184.2553 & 0.004537 \\ \hline
        WGAN GP. & CIFAR-10 & 110.9935 & 0.000946 \\ \cline{2-4}
         & AFFECTIVE & 201.5534 & 0.006299 \\ \cline{2-4}
         & A. AFFECTIVE & 181.5519 & 0.003346 \\ \hline
        WGAN GP. D. & CIFAR-10 & 116.9051 & 0.001132 \\ \cline{2-4}
         & AFFECTIVE & 202.8119 & 0.005879 \\ \cline{2-4}
         & A. AFFECTIVE & 180.0444 & 0.002024 \\ \hline
        WGAN GP. SN. & CIFAR-10 & 121.1699 & 0.001292 \\ \cline{2-4}
         & AFFECTIVE & 244.0882 & 0.010604 \\ \cline{2-4}
         & A. AFFECTIVE & 190.0199 & 0.003953 \\ \hline
    \end{tabular}}
    \label{table:scores}
\end{table}

The generation results are presented in the Tab.~\ref{table:scores}.

The first imposing conclusion is that augmenting the Affective Dataset significantly improves the performance, even up to 20\% of improvement.  

Using the dropout layers in discriminator networks also improves the performance in most cases. The best improvement was noticed for PAGAN model, up to 10\%. The worst was noticed for WGAN GP. where the improvement is marginal. 

Spectral normalization instead of batch normalization seem to provide some insignificant improvement, except for the base DCGAN model, where usage of SN totally destroyed the generated examples.

\begin{figure}[htp!]
    \centering
    [a\label{fig:dcgan_d_cifar}]{
      \includegraphics[width=0.32\textwidth]{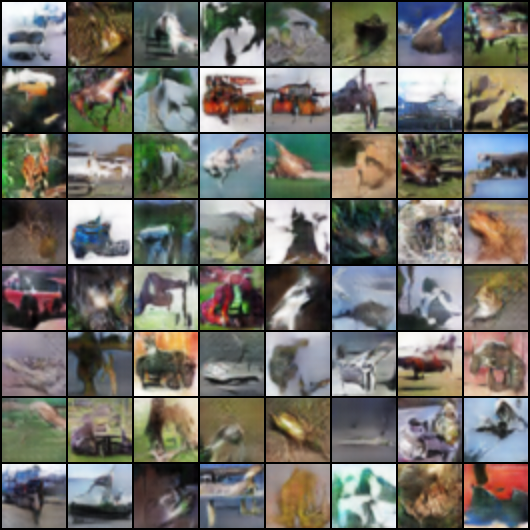}%
    }\hfil
    [b\label{fig:dcgan_d_affective}]{
      \includegraphics[width=0.32\textwidth]{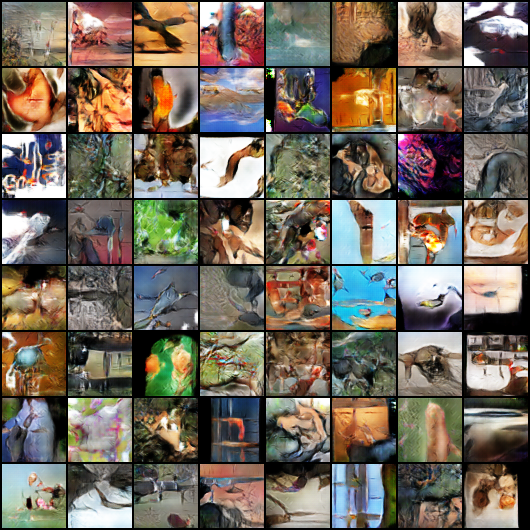}%
    }
    \caption{[a] CIFAR-10 fakes by DCGAN D., [b] A. AFFECTIVE fakes by DCGAN D.}
    \label{fig:fakes}
\end{figure}

The examples from benchmark created by DCGAN are shown in Figs.
~\ref{fig:fakes}[a] and~\ref{fig:fakes}[b]. There can be noticed that colors in affective images are more expressive than those from CIFAR-10, which indeed reflects the real situation. However, in both cases it is quite difficult to find some objects, although for CIFAR-10 some objects resembling animals or cars can be spotted. For affective images it is hard to notice any specific objects, however the generated color palettes may resemble the real data.

\begin{figure}[htp!]
    \centering
    [a\label{fig:dcgan_sn_cifar}]{%
      \includegraphics[width=0.3\textwidth]{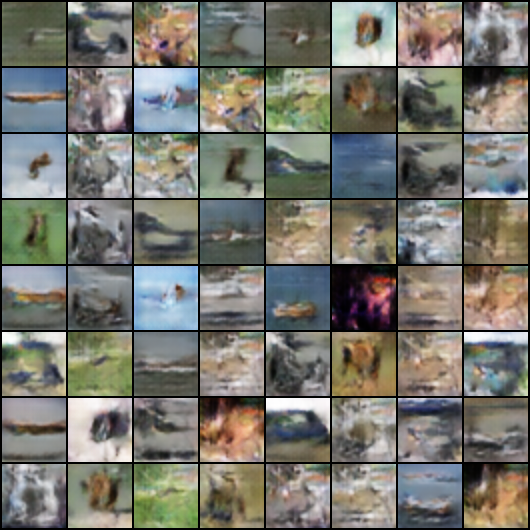}%
    }\hfil
    [b\label{fig:dcgan_sn_affective}]{%
      \includegraphics[width=0.3\textwidth]{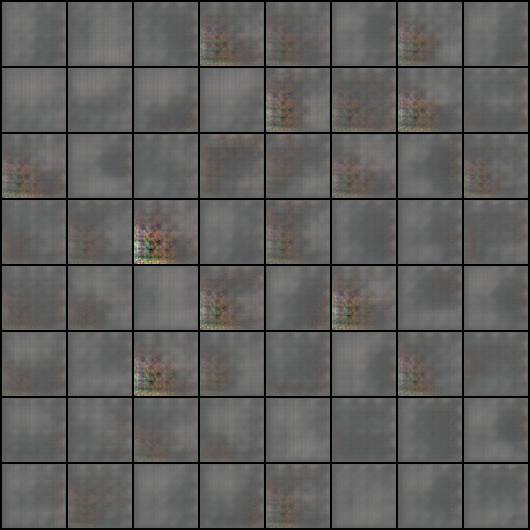}%
    }
    [c\label{fig:acgan_d_cifar}]{%
      \includegraphics[width=0.3\textwidth]{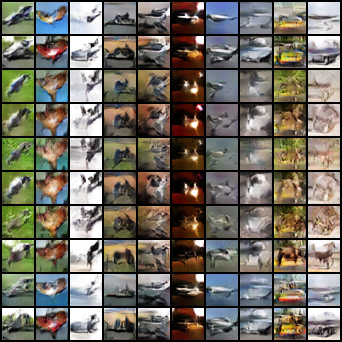}%
    }\hfil
    [d\label{fig:acgan_affective}]{%
      \includegraphics[width=0.3\textwidth]{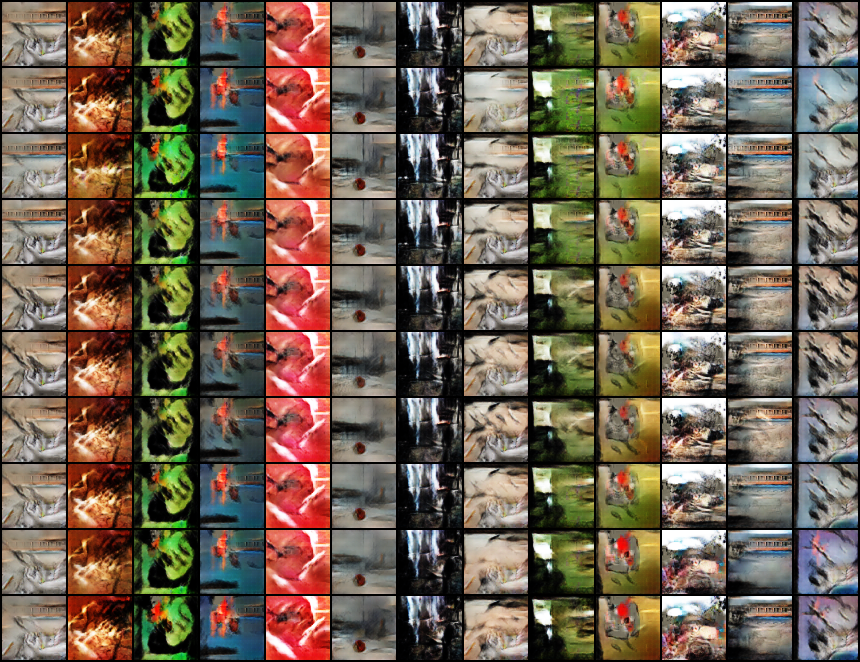}%
    }
    [e\label{fig:best_pagan}]{%
      \includegraphics[width=0.3\textwidth]{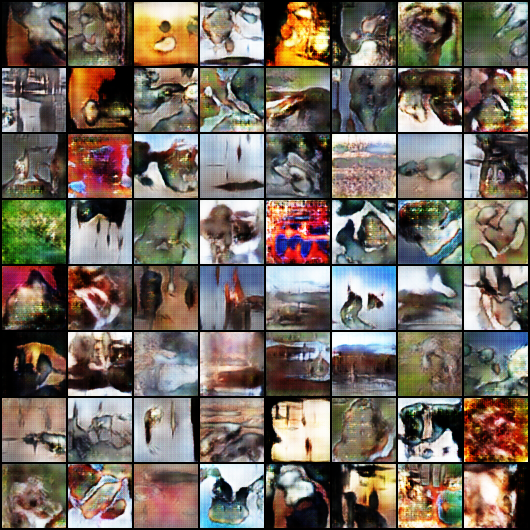}%
    }\hfil
    [f\label{fig:best_wgan}]{%
      \includegraphics[width=0.3\textwidth]{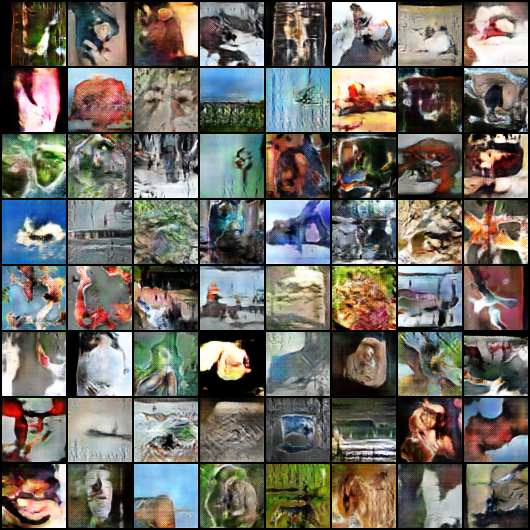}%
    }
    \caption{[a] CIFAR-10 fakes by DCGAN SN., [b] A. AFFECTIVE fakes by DCGAN SN., [c] CIFAR-10 fakes by CGAN D., [d] AFFECTIVE fakes by CGAN D., [e] AFFECTIVE fakes by PAGAN D., [f] AFFECTIVE fakes by WGAN GP. D.}
    \label{fig:fakes_big}
\end{figure}

The generated fakes presented in Figs.
~\ref{fig:fakes_big}[a] and~\ref{fig:fakes_big}[b] shows that simply replacing in Discriminator the Batch Normalization with Spectral Normalization totally destroyed the networks. The FID and KID scores presented in Tab.~\ref{table:scores} also reflect the case.

The results for the CGAN model are shown in the Figs.
~\ref{fig:fakes_big}[c] and~\ref{fig:fakes_big}[d].
For the CIFAR-10 there can be visually noticed some improvement in object detection. For Affective also appeared a motive with some objects in the center of images, however, the sharpness of colors is reduced.

The best fakes generated for the PAGAN and WGAN GP models are shown in Figs.
~\ref{fig:fakes_big}[e] and~\ref{fig:fakes_big}[f].

\begin{figure}[htp!]
    \centering
    \includegraphics[width=.7\textwidth]{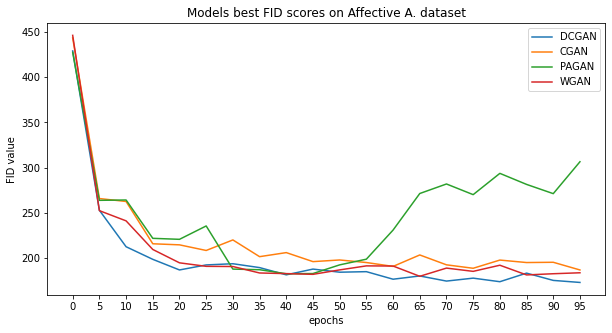}
    \caption{Models best FID scores on A. AFFECTIVE dataset}
    \label{fig:all_fid}
\end{figure}

Best FID score results for all models are juxtaposed in the Fig.~\ref{fig:all_fid}.

\subsection{Classification Results}

The results of classification task are gathered and presented in the Tab.~\ref{table:classfication_results}. There are presented accuracy scores calculated on both training and validation sets. The split between training set and validation set has been set at the 80 to 20 percents. The achieved 90\% accuracy on training set and 20\% accuracy on validation set means, that models strongly overfit. The best accuracy of 23,27\% was achieved by VGG19 on base affective dataset. Providing random assignment label for class the result would be about 7,7\%, so that means the pretrained models was able to learn a bit on affective dataset and provide better results than random. In most cases, the augmented dataset seem providing some disturbances to the classifiers inferences.

\begin{table}[htp!]
    \centering
    \caption{Accuracy of classification affective images with pretrained models}
    \begin{tabular}{|c|c|c|c|c|}
    \hline
        \textbf{Model} & \multicolumn{2}{|c|}{\textbf{AFFECTIVE}} & \multicolumn{2}{|c|}{\textbf{A. AFFECTIVE}} \\ \cline{2-5}
         & \textbf{\textit{Train}} & \textbf{\textit{Validate}} & \textbf{\textit{Train}} & \textbf{\textit{Validate}} \\ \hline
        ResNet18 & 0.9069 & 0.2029 & 0.8925 & 0.1892 \\ \hline
        ResnNet152 & 0.9143 & 0.2080 & 0.8933 & 0.2095 \\ \hline
        EfficientNet\_b7 & 0.9007 & 0.2140 & 0.8945 & 0.1950 \\ \hline
        VGG19 & 0.8525 & 0.2327 & 0.8898 & 0.2086 \\ \hline
    \end{tabular}
    \label{table:classfication_results}
\end{table}


\section{DALL-E}
\label{sec:dalle}

One would question the approach to generating affective images together with the categories of affectivity assigned to them. In this case, we redefine the task of generating affective images and categorizing synthetic images. 


In exploring the capability of generative models to produce affective images, it's essential to research the mechanisms through which these images are not only generated but also categorized based on emotional responses they elicit. The advent of sophisticated generative models like DALL-E~\cite{ramesh2021zeroshot}, which can produce high-resolution images from text descriptions, offers a compelling method for creating visual content that can induce specific emotional states in viewers.

\begin{figure}[htp!]
    \centering
    [a\label{fig:frustration}]{%
      \includegraphics[width=0.35\textwidth]{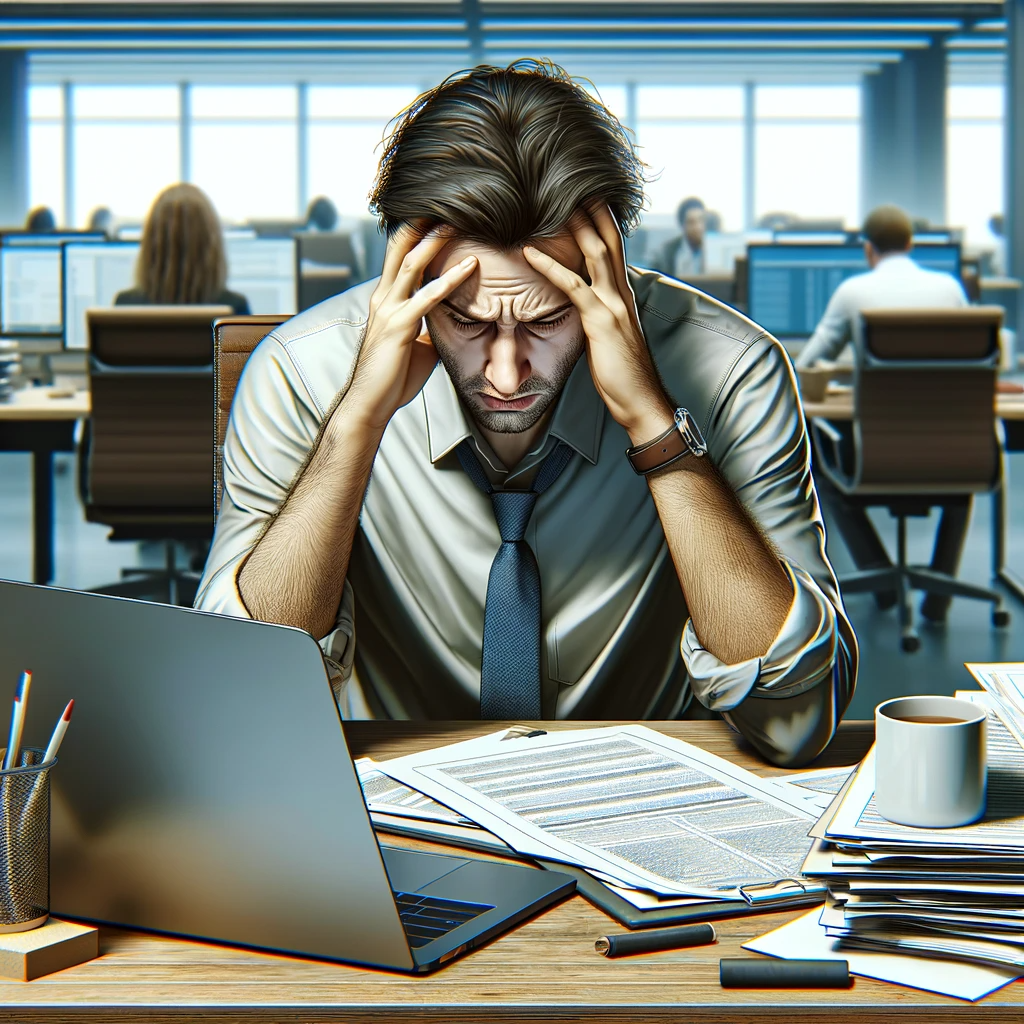}%
    }\hfil
    [b\label{fig:happinness}]{%
      \includegraphics[width=0.35\textwidth]{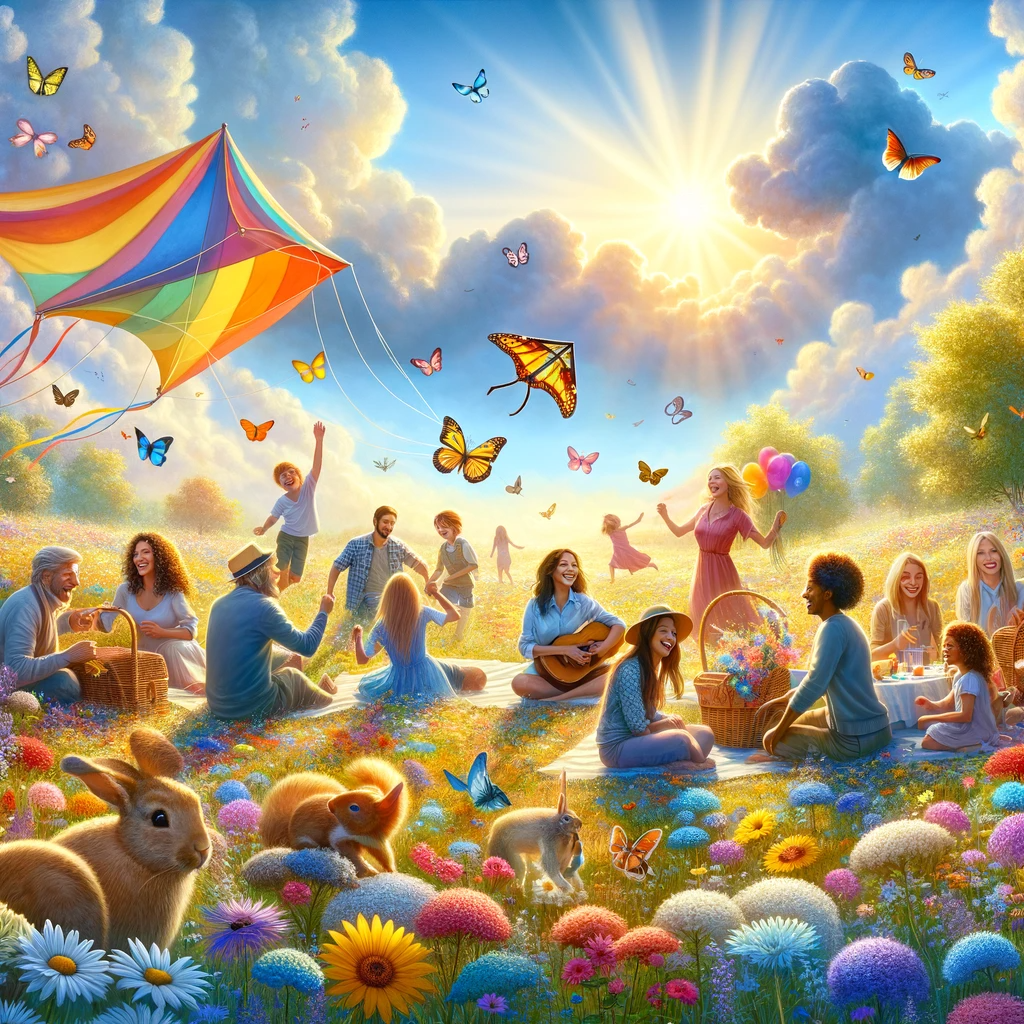}%
    }
    \caption{Generated images by DALL-E~\cite{ramesh2021zeroshot} by prompting: [a] frustration [b] happiness.}
    \label{fig:dalle}
\end{figure}

The process begins by understanding the concept of "affective images", which are images specifically designed to evoke particular emotions such as happiness, frustration, sadness, etc. By prompting models like DALL-E we can specify the desired emotional impact. Such models can tailor the visual elements in the image to align with affective categories in order to generate an image making the viewer happy or frustrated. Fig.~\ref{fig:dalle} presents two created images based on text descriptions, using prompts such as: ``generate image that will make watching person frustrated|happy|etc.''

Considering the functionality demonstrated in DALL-E, where two images are created from text prompts specifying different emotions, the model's ability to interpret and visualize such abstract concepts as emotions into tangible images underscores the advanced state of current AI technology in understanding and manipulating human emotional responses. These generated images, characterized by their high resolution and clarity can serve as tools for studying human affective responses. The quality of these images ensures that they are potent stimuli capable of reliably inducing the specified emotions.

However, the generation of affective images leads to another complex task: categorizing these images within a framework that quantifies emotional responses, often referred to as the valence-arousal space. Valence measures how positive or negative an emotion is, while arousal quantifies the intensity of the emotion. By categorizing images in this space, researchers can more systematically study how different visual elements correlate with emotional impacts.


\section{Summary and Future Work}
\label{sec:summary}


The task of generating emotionally charged images is challenging, yet feasible to some extent. Studies have shown models can capture features like color hues and saturation. Image augmentation can reduce the problem of insufficient pictures by up to 20\%, but additional datasets are needed to represent emotions like Tense, Excited, Bored, and Tired more fully.

Using dropout layers in discriminators proves effective for regularization and performance, while spectral normalization's benefits vary by model, necessitating case-by-case testing. Limiting discriminator channel increases to two during training is advisable, as exceeding this number can degrade results.

Pretrained models, including BigGAN pre-trained on ImageNet, underperformed in generating affective images, indicating the complexity of capturing emotions over object categorization. Future work should leverage conditional Deep Convolutional GANs to generate images with specific emotional labels and enhance the Affective dataset with more comprehensive human evaluations.

An augmented dataset expanded the base dataset eightfold. Further studies should explore more extensive dataset augmentation and the potential of combining promising models, like enhancing conditional GANs with additional discriminator channels or applying gradient penalty for loss improvement.

Given that transfer learning showed limited success, focusing on developing models from scratch, possibly using advanced architectures trained on an augmented dataset, may yield better outcomes. Additionally, exploring image inpainting~\cite{inpainting} as an alternative approach could offer valuable insights, requiring dataset preprocessing to generate missing image parts.

Redefinig task to only classify existing/generated images by programs like DALL-E may be the clue to get more affective images ready for future studies.

\section*{Acknowledgements}

The research for this publication has been supported by a grant from the Priority Research Area DigiWorld under the Strategic Programme Excellence Initiative at Jagiellonian University.
The research has been supported by a grant from the Faculty of Physics, Astronomy and Applied Computer Science under the Strategic Programme Excellence Initiative at Jagiellonian University.







\bibliography{bibliography.bib}

\begin{thebibliography}{33}
\expandafter\ifx\csname natexlab\endcsname\relax\def\natexlab#1{#1}\fi
\providecommand{\url}[1]{\texttt{#1}}
\providecommand{\href}[2]{#2}
\providecommand{\path}[1]{#1}
\providecommand{\DOIprefix}{doi:}
\providecommand{\ArXivprefix}{arXiv:}
\providecommand{\URLprefix}{URL: }
\providecommand{\Pubmedprefix}{pmid:}
\providecommand{\doi}[1]{\href{http://dx.doi.org/#1}{\path{#1}}}
\providecommand{\Pubmed}[1]{\href{pmid:#1}{\path{#1}}}
\providecommand{\bibinfo}[2]{#2}
\ifx\xfnm\relax \def\xfnm[#1]{\unskip,\space#1}\fi
\bibitem[{Arjovsky et~al.(2017)Arjovsky, Chintala and Bottou}]{wgan}
\bibinfo{author}{Arjovsky, M.}, \bibinfo{author}{Chintala, S.},
  \bibinfo{author}{Bottou, L.}, \bibinfo{year}{2017}.
\newblock \bibinfo{title}{Wasserstein generative adversarial networks}, in:
  \bibinfo{editor}{Precup, D.}, \bibinfo{editor}{Teh, Y.W.} (Eds.),
  \bibinfo{booktitle}{Proceedings of the 34th International Conference on
  Machine Learning, {ICML}}, \bibinfo{publisher}{{PMLR}}. pp.
  \bibinfo{pages}{214--223}.
\bibitem[{Bińkowski et~al.(2018)Bińkowski, Sutherland, Arbel and
  Gretton}]{kid}
\bibinfo{author}{Bińkowski, M.}, \bibinfo{author}{Sutherland, D.J.},
  \bibinfo{author}{Arbel, M.}, \bibinfo{author}{Gretton, A.},
  \bibinfo{year}{2018}.
\newblock \bibinfo{title}{Demystifying mmd gans}.
\newblock \URLprefix \url{https://arxiv.org/abs/1801.01401},
  \DOIprefix\doi{10.48550/ARXIV.1801.01401}.
\bibitem[{Borji(2018)}]{metrics}
\bibinfo{author}{Borji, A.}, \bibinfo{year}{2018}.
\newblock \bibinfo{title}{Pros and cons of gan evaluation measures}.
\newblock \URLprefix \url{https://arxiv.org/abs/1802.03446},
  \DOIprefix\doi{10.48550/ARXIV.1802.03446}.
\bibitem[{Borji(2021)}]{metrics2}
\bibinfo{author}{Borji, A.}, \bibinfo{year}{2021}.
\newblock \bibinfo{title}{Pros and cons of gan evaluation measures: New
  developments}.
\newblock \URLprefix \url{https://arxiv.org/abs/2103.09396},
  \DOIprefix\doi{10.48550/ARXIV.2103.09396}.
\bibitem[{Brock et~al.(2019)Brock, Donahue and Simonyan}]{biggan}
\bibinfo{author}{Brock, A.}, \bibinfo{author}{Donahue, J.},
  \bibinfo{author}{Simonyan, K.}, \bibinfo{year}{2019}.
\newblock \bibinfo{title}{Large scale {GAN} training for high fidelity natural
  image synthesis}, in: \bibinfo{booktitle}{International Conference on
  Learning Representations}.
\newblock \URLprefix \url{https://openreview.net/forum?id=B1xsqj09Fm}.
\bibitem[{Carreti{\'e} et~al.(2019)Carreti{\'e}, Tapia, L{\'o}pez-Mart{\'i}n
  and Albert}]{carretie2019emomadrid}
\bibinfo{author}{Carreti{\'e}, L.}, \bibinfo{author}{Tapia, M.},
  \bibinfo{author}{L{\'o}pez-Mart{\'i}n, S.}, \bibinfo{author}{Albert, J.},
  \bibinfo{year}{2019}.
\newblock \bibinfo{title}{Emomadrid: An emotional pictures database for affect
  research}.
\newblock \bibinfo{journal}{Motivation and Emotion} \bibinfo{volume}{43},
  \bibinfo{pages}{929--939}.
\newblock \URLprefix \url{https://doi.org/10.1007/s11031-019-09780-y},
  \DOIprefix\doi{10.1007/s11031-019-09780-y}.
\bibitem[{Dan-Glauser and Scherer(2011)}]{gaped}
\bibinfo{author}{Dan-Glauser, E.S.}, \bibinfo{author}{Scherer, K.R.},
  \bibinfo{year}{2011}.
\newblock \bibinfo{title}{The geneva affective picture database (gaped): a new
  730-picture database focusing on valence and normative significance}.
\newblock \bibinfo{journal}{Behavior Research Methods} \bibinfo{volume}{43},
  \bibinfo{pages}{468}.
\newblock \URLprefix \url{https://doi.org/10.3758/s13428-011-0064-1},
  \DOIprefix\doi{10.3758/s13428-011-0064-1}.
\bibitem[{Deng et~al.(2009)Deng, Dong, Socher, Li, Li and Fei{-}Fei}]{imagenet}
\bibinfo{author}{Deng, J.}, \bibinfo{author}{Dong, W.},
  \bibinfo{author}{Socher, R.}, \bibinfo{author}{Li, L.}, \bibinfo{author}{Li,
  K.}, \bibinfo{author}{Fei{-}Fei, L.}, \bibinfo{year}{2009}.
\newblock \bibinfo{title}{Imagenet: {A} large-scale hierarchical image
  database}, in: \bibinfo{booktitle}{2009 {IEEE} Computer Society Conference on
  Computer Vision and Pattern Recognition {(CVPR})}, \bibinfo{publisher}{{IEEE}
  Computer Society}. pp. \bibinfo{pages}{248--255}.
\newblock \DOIprefix\doi{10.1109/CVPR.2009.5206848}.
\bibitem[{Dzedzickis et~al.(2020)Dzedzickis, Kaklauskas and
  Bucinskas}]{dzedzickis2020sensors}
\bibinfo{author}{Dzedzickis, A.}, \bibinfo{author}{Kaklauskas, A.},
  \bibinfo{author}{Bucinskas, V.}, \bibinfo{year}{2020}.
\newblock \bibinfo{title}{Human emotion recognition: Review of sensors and
  methods}.
\newblock \bibinfo{journal}{Sensors} \bibinfo{volume}{20},
  \bibinfo{pages}{592}.
\newblock \URLprefix \url{https://doi.org/10.3390/s20030592},
  \DOIprefix\doi{10.3390/s20030592}.
\bibitem[{Foster(2019)}]{foster2019generative}
\bibinfo{author}{Foster, D.}, \bibinfo{year}{2019}.
\newblock \bibinfo{title}{Generative Deep Learning: Teaching Machines to Paint,
  Write, Compose, and Play}.
\newblock \bibinfo{publisher}{O'Reilly Media}.
\bibitem[{Goodfellow et~al.(2014)Goodfellow, Pouget{-}Abadie, Mirza, Xu,
  Warde{-}Farley, Ozair, Courville and Bengio}]{goodfellow-gans}
\bibinfo{author}{Goodfellow, I.J.}, \bibinfo{author}{Pouget{-}Abadie, J.},
  \bibinfo{author}{Mirza, M.}, \bibinfo{author}{Xu, B.},
  \bibinfo{author}{Warde{-}Farley, D.}, \bibinfo{author}{Ozair, S.},
  \bibinfo{author}{Courville, A.C.}, \bibinfo{author}{Bengio, Y.},
  \bibinfo{year}{2014}.
\newblock \bibinfo{title}{Generative adversarial nets}, in:
  \bibinfo{editor}{Ghahramani, Z.}, \bibinfo{editor}{Welling, M.},
  \bibinfo{editor}{Cortes, C.}, \bibinfo{editor}{Lawrence, N.D.},
  \bibinfo{editor}{Weinberger, K.Q.} (Eds.), \bibinfo{booktitle}{Advances in
  Neural Information Processing Systems 27}, pp. \bibinfo{pages}{2672--2680}.
\bibitem[{Gulrajani et~al.(2017)Gulrajani, Ahmed, Arjovsky, Dumoulin and
  Courville}]{wgan-gp}
\bibinfo{author}{Gulrajani, I.}, \bibinfo{author}{Ahmed, F.},
  \bibinfo{author}{Arjovsky, M.}, \bibinfo{author}{Dumoulin, V.},
  \bibinfo{author}{Courville, A.C.}, \bibinfo{year}{2017}.
\newblock \bibinfo{title}{Improved training of wasserstein gans}, in:
  \bibinfo{editor}{Guyon, I.}, \bibinfo{editor}{von Luxburg, U.},
  \bibinfo{editor}{Bengio, S.}, \bibinfo{editor}{Wallach, H.M.},
  \bibinfo{editor}{Fergus, R.}, \bibinfo{editor}{Vishwanathan, S.V.N.},
  \bibinfo{editor}{Garnett, R.} (Eds.), \bibinfo{booktitle}{Advances in Neural
  Information Processing Systems 30}, pp. \bibinfo{pages}{5767--5777}.
\bibitem[{He et~al.(2015)He, Zhang, Ren and Sun}]{resnet}
\bibinfo{author}{He, K.}, \bibinfo{author}{Zhang, X.}, \bibinfo{author}{Ren,
  S.}, \bibinfo{author}{Sun, J.}, \bibinfo{year}{2015}.
\newblock \bibinfo{title}{Deep residual learning for image recognition}.
\newblock \URLprefix \url{https://arxiv.org/abs/1512.03385},
  \DOIprefix\doi{10.48550/ARXIV.1512.03385}.
\bibitem[{Heusel et~al.(2018)Heusel, Ramsauer, Unterthiner, Nessler and
  Hochreiter}]{fid}
\bibinfo{author}{Heusel, M.}, \bibinfo{author}{Ramsauer, H.},
  \bibinfo{author}{Unterthiner, T.}, \bibinfo{author}{Nessler, B.},
  \bibinfo{author}{Hochreiter, S.}, \bibinfo{year}{2018}.
\newblock \bibinfo{title}{Gans trained by a two time-scale update rule converge
  to a local nash equilibrium}.
\newblock \href{http://arxiv.org/abs/1706.08500}{{\tt arXiv:1706.08500}}.
\bibitem[{Karras et~al.(2018)Karras, Laine and Aila}]{style-gan}
\bibinfo{author}{Karras, T.}, \bibinfo{author}{Laine, S.},
  \bibinfo{author}{Aila, T.}, \bibinfo{year}{2018}.
\newblock \bibinfo{title}{A style-based generator architecture for generative
  adversarial networks}.
\newblock \URLprefix \url{https://arxiv.org/abs/1812.04948},
  \DOIprefix\doi{10.48550/ARXIV.1812.04948}.
\bibitem[{Krizhevsky et~al.()Krizhevsky, Nair and Hinton}]{cifar10}
\bibinfo{author}{Krizhevsky, A.}, \bibinfo{author}{Nair, V.},
  \bibinfo{author}{Hinton, G.}, .
\newblock \bibinfo{title}{Cifar-10 (canadian institute for advanced research)}
  \URLprefix \url{http://www.cs.toronto.edu/~kriz/cifar.html}.
\bibitem[{Kurdi et~al.(2017)Kurdi, Lozano and Banaji}]{oasis}
\bibinfo{author}{Kurdi, B.}, \bibinfo{author}{Lozano, S.},
  \bibinfo{author}{Banaji, M.R.}, \bibinfo{year}{2017}.
\newblock \bibinfo{title}{Introducing the open affective standardized image set
  ({OASIS})}.
\newblock \bibinfo{journal}{Behav. Res. Methods} \bibinfo{volume}{49},
  \bibinfo{pages}{457--470}.
\bibitem[{Kutt et~al.(2022)Kutt, Dra\.{z}yk, \.Zuchowska, Szela\.zek, Bobek and
  Nalepa}]{kkt2022nature}
\bibinfo{author}{Kutt, K.}, \bibinfo{author}{Dra\.{z}yk, D.},
  \bibinfo{author}{\.Zuchowska, L.}, \bibinfo{author}{Szela\.zek, M.},
  \bibinfo{author}{Bobek, S.}, \bibinfo{author}{Nalepa, G.J.},
  \bibinfo{year}{2022}.
\newblock \bibinfo{title}{{BIRAFFE2}, a multimodal dataset for emotion-based
  personalization in rich affective game environments}.
\newblock \bibinfo{journal}{Scientific Data} \bibinfo{volume}{9},
  \bibinfo{pages}{274}.
\newblock \URLprefix \url{https://doi.org/10.1038/s41597-022-01402-6},
  \DOIprefix\doi{10.1038/s41597-022-01402-6}.
\bibitem[{Lang et~al.(2008)Lang, Bradley and Cuthbert}]{lang2008iaps}
\bibinfo{author}{Lang, P.J.}, \bibinfo{author}{Bradley, M.M.},
  \bibinfo{author}{Cuthbert, B.N.}, \bibinfo{year}{2008}.
\newblock \bibinfo{title}{International affective picture system (IAPS):
  Affective ratings of pictures and instruction manual. Technical report
  {B}-3}.
\newblock \bibinfo{type}{Technical Report}. The Center for Research in
  Psychophysiology, University of Florida. \bibinfo{address}{Gainsville, FL}.
\bibitem[{Marchewka et~al.(2014)Marchewka, {\.{Z}}urawski, Jednor{\'o}g and
  Grabowska}]{marchewka2014naps}
\bibinfo{author}{Marchewka, A.}, \bibinfo{author}{{\.{Z}}urawski, {\L}.},
  \bibinfo{author}{Jednor{\'o}g, K.}, \bibinfo{author}{Grabowska, A.},
  \bibinfo{year}{2014}.
\newblock \bibinfo{title}{The {N}encki {A}ffective {P}icture {S}ystem ({NAPS}):
  Introduction to a novel, standardized, wide-range, high-quality, realistic
  picture database}.
\newblock \bibinfo{journal}{Behavior Research Methods} \bibinfo{volume}{46},
  \bibinfo{pages}{596--610}.
\newblock \DOIprefix\doi{10.3758/s13428-013-0379-1}.
\bibitem[{Micha\l{}owski et~al.(2017)Micha\l{}owski, Dro\'zdziel, Matuszewski,
  Koziejowski, Jednor\'og and Marchewka}]{michalowski2017sfip}
\bibinfo{author}{Micha\l{}owski, J.M.}, \bibinfo{author}{Dro\'zdziel, D.},
  \bibinfo{author}{Matuszewski, J.}, \bibinfo{author}{Koziejowski, W.},
  \bibinfo{author}{Jednor\'og, K.}, \bibinfo{author}{Marchewka, A.},
  \bibinfo{year}{2017}.
\newblock \bibinfo{title}{The set of fear inducing pictures ({SFIP}):
  Development and validation in fearful and nonfearful individuals}.
\newblock \bibinfo{journal}{Behavior Research Methods} \bibinfo{volume}{49},
  \bibinfo{pages}{1407--1419}.
\newblock \URLprefix \url{https://doi.org/10.3758/s13428-016-0797-y},
  \DOIprefix\doi{10.3758/s13428-016-0797-y}.
\bibitem[{Mirza and Osindero(2014)}]{cgan}
\bibinfo{author}{Mirza, M.}, \bibinfo{author}{Osindero, S.},
  \bibinfo{year}{2014}.
\newblock \bibinfo{title}{Conditional generative adversarial nets}.
\newblock \bibinfo{journal}{CoRR} \bibinfo{volume}{abs/1411.1784}.
\newblock \URLprefix \url{http://arxiv.org/abs/1411.1784},
  \href{http://arxiv.org/abs/1411.1784}{{\tt arXiv:1411.1784}}.
\bibitem[{Odena et~al.(2017)Odena, Olah and Shlens}]{acgan}
\bibinfo{author}{Odena, A.}, \bibinfo{author}{Olah, C.},
  \bibinfo{author}{Shlens, J.}, \bibinfo{year}{2017}.
\newblock \bibinfo{title}{Conditional image synthesis with auxiliary classifier
  gans}, in: \bibinfo{editor}{Precup, D.}, \bibinfo{editor}{Teh, Y.W.} (Eds.),
  \bibinfo{booktitle}{Proceedings of the 34th International Conference on
  Machine Learning, {ICML}}, \bibinfo{publisher}{{PMLR}}. pp.
  \bibinfo{pages}{2642--2651}.
\bibitem[{Picard(1997)}]{picard1997affective}
\bibinfo{author}{Picard, R.W.}, \bibinfo{year}{1997}.
\newblock \bibinfo{title}{Affective Computing}.
\newblock \bibinfo{publisher}{MIT Press}, \bibinfo{address}{Cambridge, MA}.
\bibitem[{Radford et~al.(2016)Radford, Metz and Chintala}]{dcgan}
\bibinfo{author}{Radford, A.}, \bibinfo{author}{Metz, L.},
  \bibinfo{author}{Chintala, S.}, \bibinfo{year}{2016}.
\newblock \bibinfo{title}{Unsupervised representation learning with deep
  convolutional generative adversarial networks}, in: \bibinfo{editor}{Bengio,
  Y.}, \bibinfo{editor}{LeCun, Y.} (Eds.), \bibinfo{booktitle}{4th
  International Conference on Learning Representations {ICLR}}.
\newblock \URLprefix \url{http://arxiv.org/abs/1511.06434}.
\bibitem[{Ramesh et~al.(2021)Ramesh, Pavlov, Goh, Gray, Voss, Radford, Chen and
  Sutskever}]{ramesh2021zeroshot}
\bibinfo{author}{Ramesh, A.}, \bibinfo{author}{Pavlov, M.},
  \bibinfo{author}{Goh, G.}, \bibinfo{author}{Gray, S.}, \bibinfo{author}{Voss,
  C.}, \bibinfo{author}{Radford, A.}, \bibinfo{author}{Chen, M.},
  \bibinfo{author}{Sutskever, I.}, \bibinfo{year}{2021}.
\newblock \bibinfo{title}{Zero-shot text-to-image generation}.
\newblock \href{http://arxiv.org/abs/2102.12092}{{\tt arXiv:2102.12092}}.
\bibitem[{Salimans et~al.(2016)Salimans, Goodfellow, Zaremba, Cheung, Radford
  and Chen}]{is}
\bibinfo{author}{Salimans, T.}, \bibinfo{author}{Goodfellow, I.},
  \bibinfo{author}{Zaremba, W.}, \bibinfo{author}{Cheung, V.},
  \bibinfo{author}{Radford, A.}, \bibinfo{author}{Chen, X.},
  \bibinfo{year}{2016}.
\newblock \bibinfo{title}{Improved techniques for training gans}.
\newblock \URLprefix \url{https://arxiv.org/abs/1606.03498},
  \DOIprefix\doi{10.48550/ARXIV.1606.03498}.
\bibitem[{Shorten and Khoshgoftaar(2019)}]{data-augmentation}
\bibinfo{author}{Shorten, C.}, \bibinfo{author}{Khoshgoftaar, T.M.},
  \bibinfo{year}{2019}.
\newblock \bibinfo{title}{A survey on image data augmentation for deep
  learning}.
\newblock \bibinfo{journal}{J. Big Data} \bibinfo{volume}{6},
  \bibinfo{pages}{60}.
\newblock \DOIprefix\doi{10.1186/s40537-019-0197-0}.
\bibitem[{Simonyan and Zisserman(2014)}]{vgg}
\bibinfo{author}{Simonyan, K.}, \bibinfo{author}{Zisserman, A.},
  \bibinfo{year}{2014}.
\newblock \bibinfo{title}{Very deep convolutional networks for large-scale
  image recognition}.
\newblock \URLprefix \url{https://arxiv.org/abs/1409.1556},
  \DOIprefix\doi{10.48550/ARXIV.1409.1556}.
\bibitem[{Tan and Le(2019)}]{efficientnet}
\bibinfo{author}{Tan, M.}, \bibinfo{author}{Le, Q.V.}, \bibinfo{year}{2019}.
\newblock \bibinfo{title}{Efficientnet: Rethinking model scaling for
  convolutional neural networks} \URLprefix
  \url{https://arxiv.org/abs/1905.11946},
  \DOIprefix\doi{10.48550/ARXIV.1905.11946}.
\bibitem[{Yu et~al.(2016)Yu, Lee, Hao, Wang, He, Hu, Lai and
  Zhang}]{val-aro-diagram}
\bibinfo{author}{Yu, L.C.}, \bibinfo{author}{Lee, L.H.}, \bibinfo{author}{Hao,
  S.}, \bibinfo{author}{Wang, J.}, \bibinfo{author}{He, Y.},
  \bibinfo{author}{Hu, J.}, \bibinfo{author}{Lai, K.}, \bibinfo{author}{Zhang,
  X.}, \bibinfo{year}{2016}.
\newblock \bibinfo{title}{Building chinese affective resources in
  valence-arousal dimensions}.
\newblock \DOIprefix\doi{10.18653/v1/N16-1066}.
\bibitem[{Zeng et~al.(2020)Zeng, Lin, Yang, Zhang, Shechtman and
  Lu}]{inpainting}
\bibinfo{author}{Zeng, Y.}, \bibinfo{author}{Lin, Z.}, \bibinfo{author}{Yang,
  J.}, \bibinfo{author}{Zhang, J.}, \bibinfo{author}{Shechtman, E.},
  \bibinfo{author}{Lu, H.}, \bibinfo{year}{2020}.
\newblock \bibinfo{title}{High-resolution image inpainting with iterative
  confidence feedback and guided upsampling}.
\newblock \href{http://arxiv.org/abs/2005.11742}{{\tt arXiv:2005.11742}}.
\bibitem[{Zhang and Khoreva(2019)}]{pagan}
\bibinfo{author}{Zhang, D.}, \bibinfo{author}{Khoreva, A.},
  \bibinfo{year}{2019}.
\newblock \bibinfo{title}{Progressive augmentation of gans}, in:
  \bibinfo{editor}{Wallach, H.M.}, \bibinfo{editor}{Larochelle, H.},
  \bibinfo{editor}{Beygelzimer, A.}, \bibinfo{editor}{d'Alch{\'{e}}{-}Buc, F.},
  \bibinfo{editor}{Fox, E.B.}, \bibinfo{editor}{Garnett, R.} (Eds.),
  \bibinfo{booktitle}{Advances in Neural Information Processing Systems 32:
  Annual Conference on Neural Information Processing Systems 2019, NeurIPS
  2019, December 8-14, 2019, Vancouver, BC, Canada}, pp.
  \bibinfo{pages}{6246--6256}.

\end{thebibliography}
\bibliographystyle{elsarticle-harv}







\clearpage

\normalMode







\end{document}